\pdfoutput=1

\documentclass[11pt]{article}

\usepackage[final]{acl2025}

\usepackage{times}
\usepackage{latexsym}

\usepackage[T1]{fontenc}

\usepackage[utf8]{inputenc}

\usepackage{microtype}

\usepackage{inconsolata}

\usepackage{amsmath}
\usepackage{amssymb}
\usepackage{mathtools}
\usepackage{amsthm}
\usepackage{float}
\usepackage{multirow}
\usepackage{graphicx}
\usepackage{subfigure}
\usepackage{algorithm}
\usepackage{algpseudocode}
\usepackage{footmisc}
\usepackage[ruled,vlined,algo2e,linesnumbered]{algorithm2e}

\title{C2T: A Classifier-Based Tree Construction Method in Speculative Decoding}

\author{
 \textbf{Feiye Huo\textsuperscript{1,2,*}},
 \textbf{Jianchao Tan\textsuperscript{2,*}},
 \textbf{Kefeng Zhang\textsuperscript{2}},
 \textbf{Xunliang Cai\textsuperscript{2}},
 \textbf{Shengli Sun\textsuperscript{1,+}}
\\
\\
 \textsuperscript{1}Peking University,
 \textsuperscript{2}Meituan
\\
 \small{
   \textbf{Correspondence:} phiyeh@stu.pku.edu.cn, 
   \{tanjianchao02, zhangkefeng, caixunliang\}@meituan.com, sunshengli@pku.edu.cn
 }
}

\begin{document}
\maketitle
\begin{figure*}[htbp]
\centering
\subfigure[]{
\begin{minipage}[t]{0.45\textwidth}
\centering
\includegraphics[width=\textwidth]{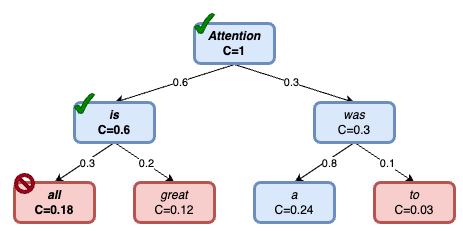}
\label{fig:shortage1}
\end{minipage}
}
\subfigure[]{
\begin{minipage}[t]{0.45\textwidth}
\centering
\includegraphics[width=\textwidth]{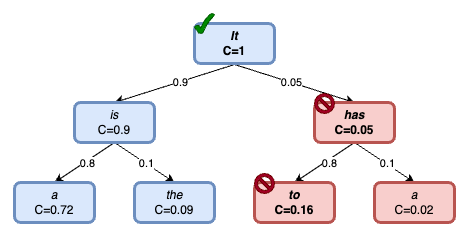}
\label{fig:shortage2}
\end{minipage}
}
\caption{Two illustrations of EAGLE-2 for verifying with 4 candidate tokens. Blue represents the chosen candidate tokens, red represents the tokens that were not chosen, bold text represents the correct answers, numbers on the arrows represent the generation probabilities, and C represents confidence, which in EAGLE-2 refers to joint probability.}
\label{fig:shortage}
\end{figure*}
\begin{abstract}
The growing scale of Large Language Models (LLMs) has exacerbated inference latency and computational costs. Speculative decoding methods, which aim to mitigate these issues, often face inefficiencies in the construction of token trees and the verification of candidate tokens. Existing strategies, including chain mode, static tree, and dynamic tree approaches, have limitations in accurately preparing candidate token trees for verification. We propose a novel method named \textbf{C2T} that adopts a lightweight classifier to generate and prune token trees dynamically. Our classifier considers additional feature variables beyond the commonly used joint probability to predict the confidence score for each draft token to determine whether it is the candidate token for verification. This method outperforms state-of-the-art (SOTA) methods such as EAGLE-2 on multiple benchmarks, by reducing the total number of candidate tokens by 25\%, while maintaining or even improving the acceptance length.
\end{abstract}

\section{Introduction}
\label{introduction}

Large Language Models (LLMs) \citet{achiam2023gpt, touvron2023llama, chiang2023vicuna} have shown remarkable abilities in various fields, but face significant bottlenecks in autoregressive token generation due to high memory bandwidth demands and underutilized GPU resources \citet{patterson2004latency, shazeer2019fast}, as each token requires access to all model parameters \citet{radford2019language, brown2020language}. To address this issue, Speculative Decoding (SD) \citet{chen2023accelerating, leviathan2023fast} has been developed, which quickly generates multiple draft tokens and verifies them all at once using the target model to maximize GPU computational capacity, and it has been applied in the latest influential LLMs \cite{liu2024deepseek}.

Vanilla SD employs a chain structure for the draft tokens, and the verification process follows a topological order \citet{chen2023accelerating, leviathan2023fast}. If a token is rejected, all subsequent tokens are also discarded. To overcome inefficiency, tree-structured draft tokens have been proposed \citet{miao2024specinfer, sun2024spectr}, which integrate multiple chains. Static tree methods, such as EAGLE-1 \citet{li2024eagle} and Medusa \citet{cai2024medusa}, use preset tree structures that bias the sampling rate of specific positions \citet{chen2024sequoia}, while dynamic methods, such as EAGLE-2 \citet{li2024eagle2}, rely on contextual information.

For dynamic tree methods, most are designed to build and prune trees on the basis of confidence scores. The most straightforward approach is to use the joint probability as confidence \citet{li2024eagle2, wang2024opttreespeculativedecodingadaptive, brown2024dynamic, qin2024dynamic}. However, directly using joint probability as confidence is not enough for complex situations, leading to misjudgments. As shown in Figure \ref{fig:shortage}, which will be explained in detail in \ref{sec:motivation}. From the perspective of tree data structure, whether a token is accepted or not, depends not only on its own property but also other nodes' properties. This means that incorporating more variables together is necessary in the confidence score calculation. Therefore, we propose a tree construction method based on a designed tiny classifier to perform this calculation. Our contributions are summarized below:

\begin{itemize}
    \item We propose a classifier-based method named \textbf{C2T} for the dynamic construction of the token tree. And it can be easily integrated into any confidence-based SD system.
    
    \item Our classifier demonstrates strong transferability across different datasets and within the same model family, thus being plug-and-play.
    
    \item Compared to the SOTA method EAGLE-2, C2T reduces the number of candidate tokens by $25\%$ while maintaining or improving the acceptance length on multiple benchmarks.
\end{itemize}

\section{Background}

\subsection{Speculative Decoding}

Speculative decoding (SD) \citet{chen2023accelerating, leviathan2023fast} is an algorithm designed to speed up model reasoning by leveraging the parallel computing capabilities of attention mechanisms. It consists of two main stages. The first stage is the draft phase, where a smaller model, known as the draft model $M_d$, generates draft tokens. The second stage is verification, where the draft tokens are verified all at once by the larger model, known as the target model $M_t$. At the same time, we call the draft tokens chosen to be verified as candidate tokens.

\subsection{Tree Attention and EAGLE-2}

Vanilla SD uses a chain-like structure, where if a token is rejected, all subsequent tokens are also discarded. SpecInfer's \citet{miao2024specinfer} Tree Attention achieves the integration of multiple speculations at a minimal cost and has been widely adopted in other SD methods \citet{he2023rest, sun2024spectr, cai2024medusa, li2024eagle, chen2024sequoia, svirschevski2024specexec}. In the pioneer SD works, the token tree was static, with a preset tree shape, and the draft model generated tokens layer by layer and filled in the corresponding positions. This static method is undoubtedly rough, so various heuristic dynamic generation methods appeared later \citet{li2024eagle2, wang2024opttreespeculativedecodingadaptive, brown2024dynamic, qin2024dynamic, huang2024specdec++}. Several trainable dynamic methods have been also proposed \citet{mamou2024dynamic, huang2024specdec++, zhang2024adaeagle}, but due to design redundancies, they have only been applied to early stopping in chain-like drafting, which significantly limits their applicability. 

The current SOTA dynamic tree construction method EAGLE-2 \citet{li2024eagle2} introduces the joint probability of each node as contextual information, and divides the sampling process into two stages: expand and rerank. The former is to build the tree and the latter is for post-pruning. We can denote the token tree after expansion as $T_1$, and the token tree reranked as $T_2$. There are three important parameters in EAGLE-2: 
\begin{enumerate}
    \item \textbf{$K$}: During expanding, each token in current layer generates Top$K$ tokens. Then, among all the generated tokens, the Top$K$ tokens are selected to generate the next tree layer. 
    \item \textbf{$d_{max}$}: The rounds of the drafting in the expansion phase.
    \item \textbf{$N$}: Use joint probability to rerank all nodes in $T_1$, and then take the Top$N$ nodes as $T_2$. Since the joint probability of the parent node is always greater than that of the child nodes, $T_2$ must be a valid subtree of $T_1$. Therefore, Top$N$ determines the size of $T_2$, which directly determines the number of candidate tokens.
\end{enumerate}

However, to achieve the benefits claimed by EAGLE-2, the target model must verify nearly twice as many tokens, which is not a fair comparison. When aligning for the size of the token tree, the benefit of EAGLE-2 in terms of accept length is only 11\%. Moreover, due to the additional latency introduced by the dynamic strategy, the wall-clock time of EAGLE-2 is more. Details in Appendix \ref{apx:eagle1 and eagle2}. 

\begin{figure*}[htbp]
\centering
\subfigure[Average Probability Heatmap]{
\begin{minipage}[t]{0.315\textwidth}
\centering
\includegraphics[width=\textwidth]{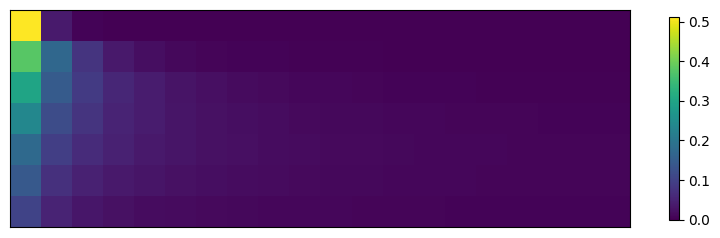}
\label{fig:e10}
\end{minipage}
}
\subfigure[Accept Rate Heatmap]{
\begin{minipage}[t]{0.315\textwidth}
\centering
\includegraphics[width=\textwidth]{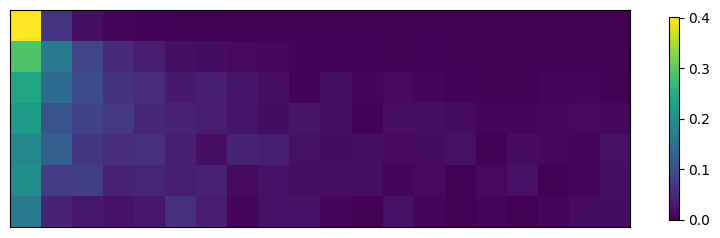}
\label{fig:e11}
\end{minipage}
}
\subfigure[Bias Heatmap]{
\begin{minipage}[t]{0.315\textwidth}
\centering
\includegraphics[width=\textwidth]{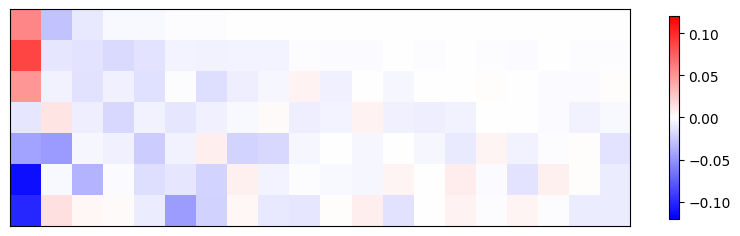}
\label{fig:e12}
\end{minipage}
}
\centering
\caption{The coordinates in the three heatmaps have the same meanings: The y-axis represents the entropy value interval in which the current probability distribution lies. From top to bottom, these intervals are 0$\sim$1, 1$\sim$2, 2$\sim$3, 3$\sim$4, 4$\sim$5, 5$\sim$6, and $>$6. The x-axis shows the top 20 probabilities within the distribution, decreasing from left to right. Each square in the heatmap indicates the value corresponding to the probability rank within the respective entropy interval. In Figure \ref{fig:e10}, the value represents the average probability at each position, smoothed using a logarithm. In Figure \ref{fig:e11}, the value represents the accept rate at each position, also smoothed with a logarithm. In Figure \ref{fig:e12}, the value shows the bias between probability and accept rate, with red indicating probabilities higher than accept rates and blue indicating the opposite.}
\label{fig:e1}
\end{figure*}

\section{Motivation}
\label{sec:motivation}
As shown in Figure \ref{fig:e1}, we use the EAGLE-2's LLaMA-2 7B model pair to reason on the MT-bench. We performed a single forward to calculate the entropy and divided it into seven intervals. We then computed the average probability of different ranks within these intervals to create the heatmap in Figure \ref{fig:e10} which shows that lower entropy corresponds to more concentrated distributions.

We also tested the acceptance rate of tokens with different ranks in these intervals and generated the heatmap in Figure \ref{fig:e11}. The similarity between \ref{fig:e10} and \ref{fig:e11} confirms the correlation between probability and acceptance rate.

However, Figure \ref{fig:e11} is less stable than Figure \ref{fig:e10}, indicating that there is a discrepancy. To quantify this, we subtracted the data matrix of Figure \ref{fig:e10} from that of Figure \ref{fig:e11} to create Figure \ref{fig:e12}, which shows the bias between probability and acceptance rate. 

As shown in Figure \ref{fig:e12}, there are two main deviations between probability and acceptance rate: 
\begin{itemize}
    \item When the entropy is high, the node with the highest probability in the distribution is underestimated in terms of probability;

    \item When the entropy is low, the node with the highest probability in the distribution is overestimated in terms of probability.
\end{itemize}
 
These two deviations may lead to misjudgments as shown in Figure \ref{fig:shortage1}. For the "all" and "great", although the probabilities of their parent are greater than those of the parent of "a" and "to", due to their higher entropy in the probability distribution, the normalized generation probabilities are not high. As a result, they are outperformed in joint probability by the "a" from a distribution with lower entropy.

In addition, due to the nature of the token tree, if a token is rejected, all its subsequent child nodes will also be rejected. Therefore, the acceptance rate is a value that decreases with depth. Nodes at shallower levels should be given greater confidence. If only joint probability is considered, the depth factor will become inactive in some cases. As shown in Figure \ref{fig:shortage2}, although the "the" and the "has" are both the smaller ones in their respective distributions and the "has" is shallower, the joint probability of the former is still greater than that of the latter because the probability of the former's parent node is too large (our previous experiments can also prove that in distributions with lower entropy, the node with the highest probability is overestimated). Consequently, in subsequent recalls, the "has" node is also unlikely to be selected, even though it is on the second level.

Therefore, directly using joint probability as confidence cannot handle complex situations and will inevitably lead to misjudgments. This necessitates that we consider more features like entropy and depth when designing the confidence function. However, introducing more features makes the function design more challenging, and low-dimensional functions are not robust. Naturally, we thought of introducing a learnable neural network to fit an efficient confidence function for us. To this end, we propose an efficient method using \textbf{C}lassifier \textbf{to} build \textbf{T}rees named \textbf{C2T} in speculative decoding. Please refer to Appendix \ref{apx:confidence} for how the classifier addresses the two situations mentioned above.

\begin{figure*}[ht]
\centering
\includegraphics[width=0.9\textwidth]{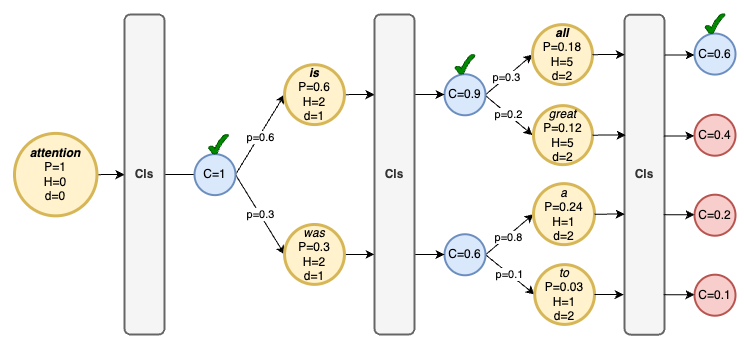}
\caption{This figure provides further details on C2T. The classifier is a two-layer FFN, represented by the rounded rectangle labeled "Cls". It uses the joint probability $P$, the entropy $H$, and its depth $d$ as features, which are depicted as the larger yellow circles, where bold text represents the correct answers. The classifier outputs a logit as the confidence score $C$, shown as the smaller circle, where blue represents the candidate tokens, and red represents the tokens that were not recalled. Then a threshold $\beta$ = 0.5 and Top$K$ = 2 is used for pre-pruning. Tokens above $\beta$ will be used as input for the next round of draft model $M_d$ to generate the next tree layer. The features for the next layer of tokens can be obtained from the generation probability $p$ output by the draft model $M_d$.}
\label{fig:detail}
\end{figure*}

\section{Methodology}

Our primary goal is to maintain the accept length with fewer candidate tokens using a lightweight classifier. To enhance its versatility, the training features should be based solely on the token tree's properties, without relying on model or dataset information. This classifier will serve as a predictor for pre-pruning during token tree construction, helping create a concise tree efficiently.

\subsection{Classifier}

The ultimate goal of our method is to reduce latency, so the classifier must be as lightweight as possible. At the same time, we aim for a plug-and-play solution where a model, trained on one dataset, remains effective when applied to others, avoiding features that are strongly tied to specific datasets or models, such as hidden states. Instead, we should use only the mathematical properties of nodes as features for the classifier. Additionally, these features should be easy to compute and minimal in number.

We adopt a two-layer Feed-Forward Network (FFN) as the classifier to achieve our goal, using ReLU as the activation function and sigmoid for normalizing the final logits. This classifier is extremely simple, with minimal parameters and computational load. For features, we start with the joint probability of each node, as methods like EAGLE-2 have shown their effectiveness as a confidence measure. Next, we select the entropy of the probability distribution in which each node resides as the third feature, as lower entropy indicates a higher acceptance rate. Finally, we include the depth of each node, since shallower nodes have a higher likelihood of acceptance. To simplify calculations, we only consider the top 1000 probabilities of each node's child nodes when calculating entropy. This simplification is necessary and almost lossless, and the detailed proof can be found in Appendix \ref{apx:proof}. By selecting the top 1000 based on probability, we can almost certainly include the topK of the confidence scores (K is typically much smaller than 1000). 

The introduction of entropy and depth is not only because they are related to the acceptance rate, but also because the dimensionality-increased node information is more conducive to solving the problems raised by only using joint probability in the motivation. More details in Appendix \ref{apx:confidence}.

With these features, a complete classifier structure is obtained. More details of training settings are shown in Appendix \ref{apx:classifier}.

\subsection{Tree Construction}

C2T is a pre-pruning approach, which can be divided into two steps: the first pruning based on confidence and the second pruning based on topK.

\subsubsection{First Pruning based on Confidence}

As shown in Figure \ref{fig:detail}, after obtaining the classifier, we can use it to build the token tree. The tree construction based on the classifier is a layer-by-layer construction process. The draft model can obtain the generation probability of each node during each forward pass. Let a node be $i$, and the generation probability of this node is $p_i$, then the joint probability of this node is denoted as 

\begin{equation}
    P_i = \prod_{j \in Path(root,i)}p_j
\end{equation}

Where the $Path(root, t_i)$ represents the set of all nodes on the path from the root node to node $i$. 

At this point, we can also determine the entropy of node $i$'s probability distribution.

\begin{align}
    H_{i} & = - \sum_{j \in S(i)} p_j \log p_j \\
          & \approx - \sum_{j \in S_{1000}(i)} p_j \log p_j
\end{align}

Where $S(i)$ represents the probability distribution of $i$, and $S_{1000}(i)$ represents the set of nodes with top 1000 probabilities. 

We can also easily record the depth of node $d_i$. 

After obtaining these three features, we can use the trained classifier to obtain the confidence of node $i$, denoted as

\begin{equation}
    C_i = F(P_i, H_i, d_i)
\end{equation}

\begin{algorithm2e}[ht]
\SetAlgoLined
\caption{C2T}
\label{alg:C2T}
\KwIn{draft model $M_d$, root node $r$, maximum depth $d_{max}$, threshold $\beta$}
\KwResult{token tree $T$}
current depth $d \gets 0$\\
confidence set $C \gets \{C_r=1\}$\\
token tree $T \gets \{r\}$\\
current tree layer node set $N \gets \{r\}$\\
\While{$d < d_{max}$ and exists $C_i > \beta$ in $C$}{
     $N \gets topK(M_d(N))$ \\
     $C \gets \{\}$ \\
    \For{$i$ in $N$}{
         $P_i \gets \prod_{j \in Path(root,i)}p_j$\\
         $d_i \gets d$\\
         $H_i \gets - \sum_{j \in S_{1000}(i-1)} p_j \log p_j$\\
         $C_i \gets F(P_i, H_i, d_i)$\\
         Append $C_i$ to $C$\\
        \If{$C_i < \beta$}{
             Remove node $i$ from $N$\\
        }
    }
     $d \gets d + 1$\\
     extend $N$ or $topK(N)$ to $T$\\
}
\textbf{return} $T$\\
\end{algorithm2e}

Where $F(*)$ represents the classifier's output. Since $C_i$ is normalized, we set a threshold $\beta$ between 0 and 1 to determine which tokens participate in generating the next tree layer. This screening-generation process repeats at each tree layer until the current depth $d_i$ reaches the maximum depth $d_{max}$ we set, or until no nodes in the current tree layer have confidence greater than $\beta$.

\subsubsection{Second Pruning based on Topk}

In our experiments, setting $\beta$ appropriately ensures that the classifier does not qualify too many tokens at each tree layer, making the naive generation method sufficient. However, in practical applications, we often need to set Top$K$ to manage tree generation. Top$K$ serves two main purposes:

\begin{enumerate}
    \item To reduce the classifier's computational cost, we calculate confidence only for the Top$K$ tokens with the highest generation probability. 
    \item To prevent excessive tree expansion, we limit the number of tokens participating in the next tree layer's generation. After identifying the tokens that pass the classifier test in the current tree layer, we select only the Top$K$ tokens with the highest confidence as the final candidates.
\end{enumerate}

Appendix \ref{sec:topk} shows that when $\beta$ is sufficiently large, the use of Top$K$ does not impact the method's effectiveness. However, when $\beta$ is less strict, Top$K$ helps prevent the number of candidate tokens from becoming too large, though it may also limit the algorithm's maximum performance. Generally, the first strategy is essential for reducing classifier costs proven in Appendix \ref{apx:proof} \textit{\textbf{Proof-2}}, while the second can be applied as needed, with the size of Top$K$ adjusted based on GPU's capabilities or omitted if unnecessary. 

\section{Experiments}

\textbf{Models:} We used LLaMA-2-Chat 7B, 13B, 70B \citet{touvron2023llama} and Vicuna 7B, 13B, 33B \citet{chiang2023vicuna} as target model $M_t$, and the corresponding draft model $M_d$ is from EAGLE \citet{li2024eagle}.

\textbf{Tasks:} To compare with EAGLE-2 \citet{li2024eagle2}, we aligned with it on the dataset. For tasks such as multi-round dialogue, code generation, mathematical reasoning, instruction following, summarization, and Q\&A, we selected the MT-bench \citet{zheng2023judging}, HumanEval \citet{chen2021evaluating}, GSM8K \citet{cobbe2021training}, Alpaca \citet{taori2023stanford}, CNN/Daily Mail \citet{nallapati2016abstractive}, and Natural Questions \citet{kwiatkowski2019natural}, respectively.

\textbf{Metrics:} The ultimate goal of this work is to let the $M_t$ verify as few candidate tokens as possible while maintaining the hit rate unchanged or even better, so we mainly focus on the following two device-independent indicators:

\begin{itemize}
    \item \textbf{The number of candidate tokens $\gamma$:} The total number of tokens verified by $M_t$.
    
    \item \textbf{Accept length $\tau$:} The average length accepted by $M_t$ for each generation, and this indicator in this paper does not include the initial token generated by the target model itself, so it needs to be added 1 compared to some papers.
\end{itemize}

In this paper, the experiment with the LLaMA-2 7B model focuses on precision rather than inference time, as the method's main advantage is obtaining a more accurate token tree at minimal cost. For powerful GPUs and lightweight LLMs, if the GPU's parallel computing capability threshold is not reached, this optimization may not improve time performance. So we will demonstrate the efficiency advantage of this method on the LLaMA-2 70B model. When analyzing inference time, there will be the following two indicators:

\begin{itemize}

    \item \textbf{Draft time:} The total time measured from the completion of the last verification to the generation of the next verification input on benchmark data.

    \item \textbf{Verify time:} The total time used for verification on benchmark data.
\end{itemize}

\textbf{Comparison:} This work will mainly compare with the SOTA dynamic strategy, EAGLE-2 \citet{li2024eagle2}. The $N$ will be our variable to control $\gamma$ and $\tau$. To make the comparison fairer, we will set EAGLE-2's $K = 15$ and $d_{max}=10$. This parameter configuration nearly reaches the limit of EAGLE-2's $M_d$ capabilities.

\begin{figure*}[htbp]
\centering
\begin{minipage}{0.33\textwidth}
\includegraphics[width=\textwidth]{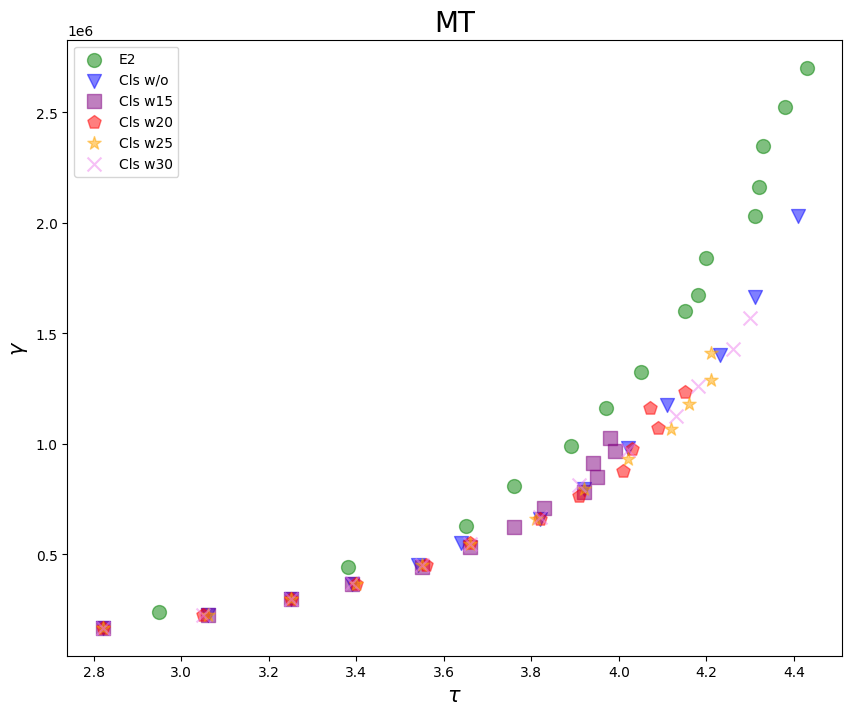}
\end{minipage}%
\begin{minipage}{0.33\textwidth}
\includegraphics[width=\textwidth]{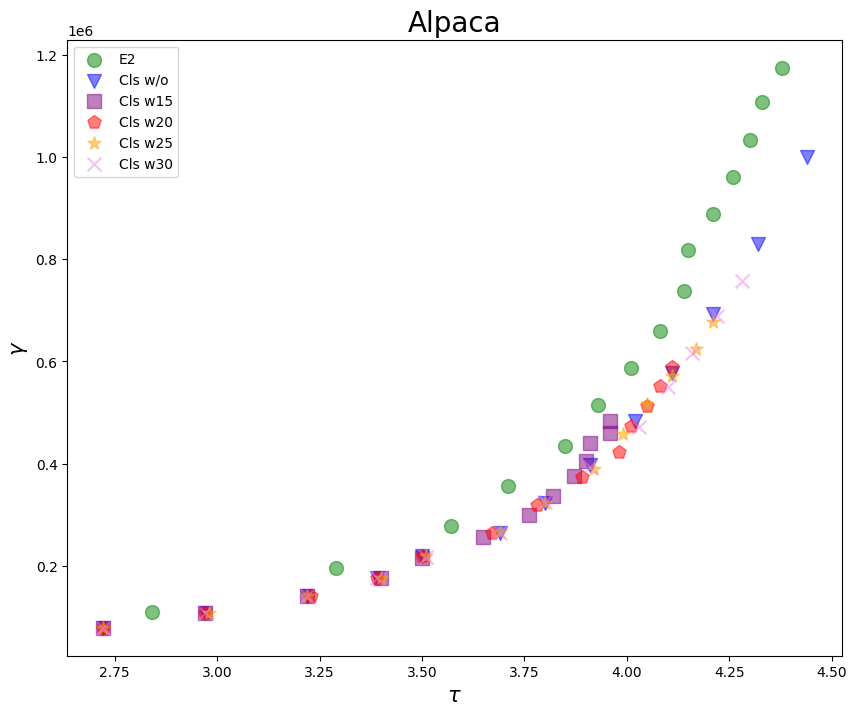}
\end{minipage}%
\begin{minipage}{0.33\textwidth}
\includegraphics[width=\textwidth]{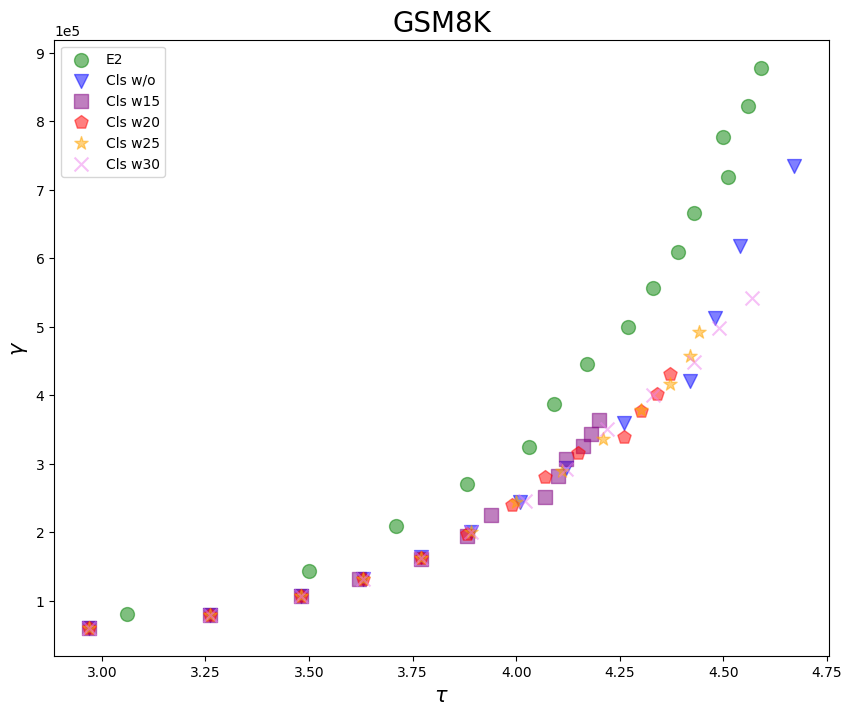}
\end{minipage}

\begin{minipage}{0.33\textwidth}
\includegraphics[width=\textwidth]{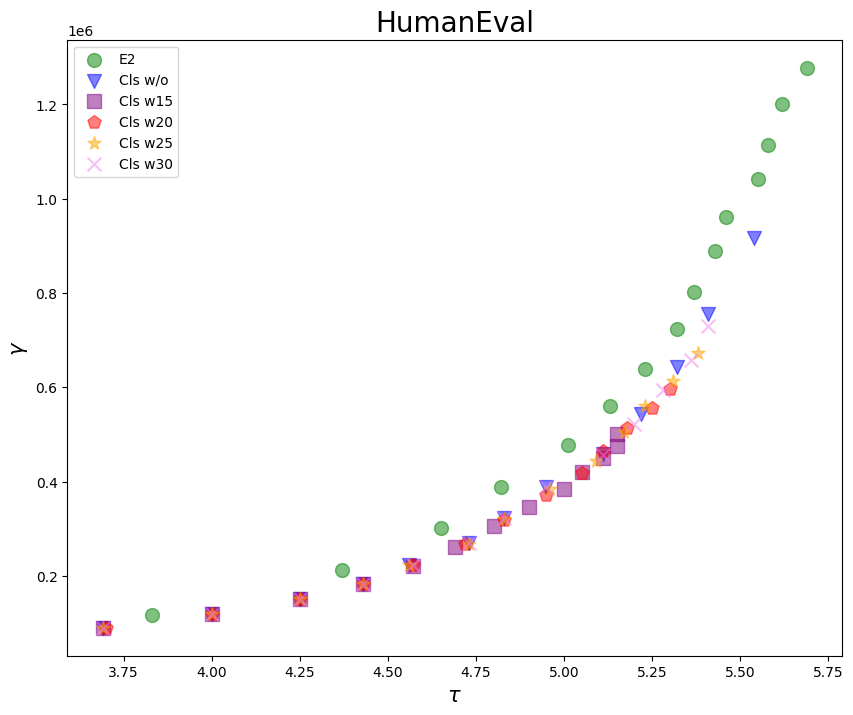}
\end{minipage}%
\begin{minipage}{0.33\textwidth}
\includegraphics[width=\textwidth]{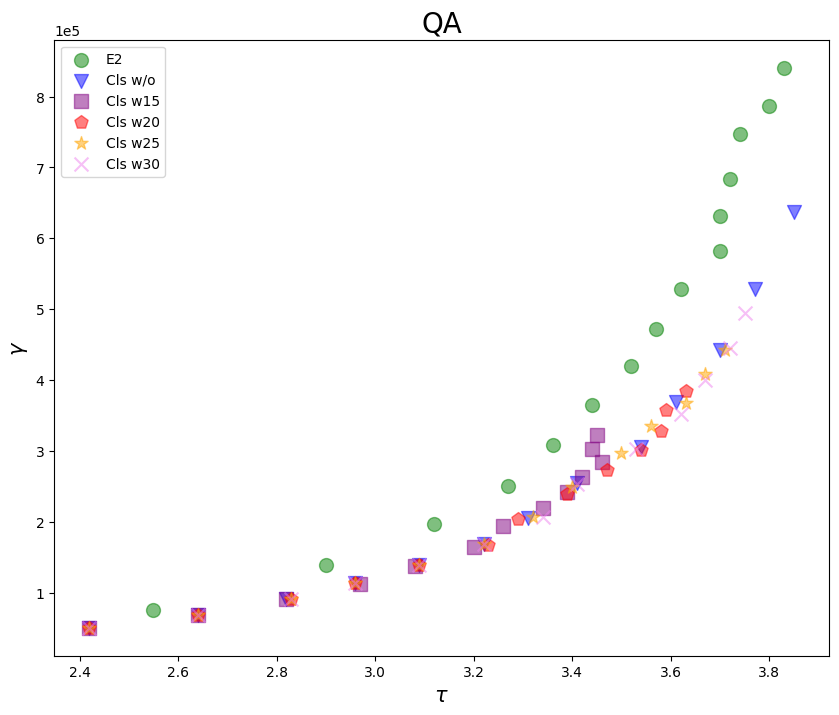}
\end{minipage}%
\begin{minipage}{0.33\textwidth}
\includegraphics[width=\textwidth]{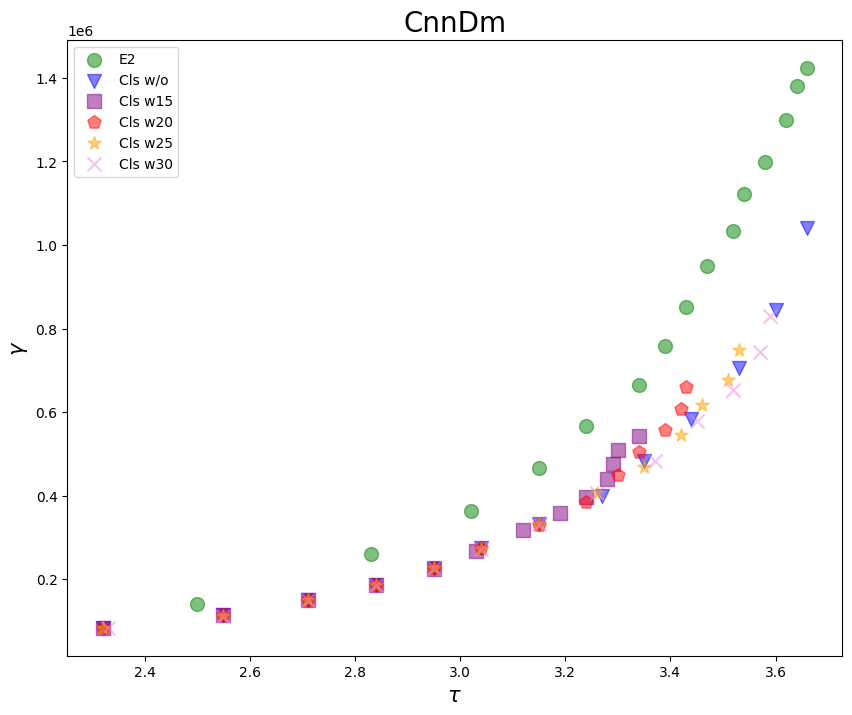}
\end{minipage}
\caption{The scatter plot uses the LLaMA-2 7B model, with the acceptance length $\tau$ on the x-axis and the number of candidate tokens $\gamma$ on the y-axis. Cls represents our classifier-based method C2T, E2 represents EAGLE-2, w/o represents not using topK secondary pruning, and w$K$ represents the use of Top$K$ secondary pruning with $K$ values of 15, 20, 25, and 30.}
\label{fig:e3}
\end{figure*}

\subsection{Feature Ablation}
We conducted ablation studies on three features: joint probability, entropy, and depth. The model trained with all three features served as the baseline, while other combinations were used as control groups. As shown in Table \ref{tbl:feature_ablation}, the joint probability is the most critical factor for performance improvement. In contrast, entropy and depth alone lead to significant performance degradation. Entropy and depth primarily serve as corrective factors, while joint probability is essential for confidence scoring. Combining joint probability with either entropy or depth slightly outperforms EAGLE-2. However, using all three features together significantly enhances performance, demonstrating their strong complementarity.

\begin{table}[H]
\centering
\renewcommand{\arraystretch}{1.25}
\begin{tabular}{c c c c c}
\hline
\textbf{Method} & \textbf{Setting} & \textbf{Feature} & \textbf{$\tau$} & \textbf{$\gamma$} \\
\hline
E-2 & $N=100$ & $/$ & 3.89 & 1.0M \\
\hline
\multirow{4}{*}{C2T} & $\beta=0.42$ & P+H & 3.92 & 0.9M \\
                     & $\beta=0.7$ & P+d & 3.92 & 0.9M \\
                     & $\beta=0.5$ & H+d & 0.98 & 3.3M \\
                     & $\beta=0.5$ & P+H+d & 3.92 & 0.7M \\
\hline
\end{tabular}
\renewcommand{\arraystretch}{1}
\caption{Regarding the ablation experiments of the classifier features, E-2 represents EAGLE-2. For both methods, we set the default $d_{max}$ to 10 and topK to 15. All experiments were conducted using LLaMA-2 7B on MT-bench. In the features, P stands for joint probability, H for entropy, and d for depth. P+H+d indicates training the classifier using joint probability, entropy, and depth as features, and so on. M represents a million. $\tau$ represents accept length and $\gamma$ represents the number of candidate tokens.}
\label{tbl:feature_ablation}
\end{table}

\subsection{Dataset Transferability}
\label{sec:data trans}

EAGLE-2, being a post-pruning method, directly controls $\gamma$ by adjusting Top$N$. In contrast, our pre-pruning approach controls $\gamma$ by setting a pruning threshold $\beta$ during tree generation. Thus, a direct step-by-step comparison is not feasible. Instead, we compare the precision of both methods using a scatter plot of $\tau$ versus $\gamma$. To further validate transferability and data independence, we transferred the classifier's parameters to other datasets without fine-tuning.

As shown in Figure \ref{fig:e3}, C2T consistently achieves lower $\gamma$ for similar $\tau$ compared to EAGLE-2, indicating superior precision. Moreover, C2T's curve is more gradual, showing its increasing advantage as $\tau$ grows. Cross-validation is in Appendix \ref{apx:cross validation}.

\begin{table}[H]
\centering
\renewcommand{\arraystretch}{1.2}
\begin{tabular}{c c c c c}
\hline
\textbf{Model} & \textbf{Method} & \textbf{Setting} & \textbf{$\tau$} & \textbf{$\gamma$} \\
\hline
\multirow{2}{*}{L2 7B} & E-2 & $N=290$ & 4.40 & 2.6M \\
                       & C2T & $\beta=0.3$ & 4.41 & 2.0M \\
\hline
\multirow{2}{*}{L2 13B} & E-2 & $N=300$ & 4.58 & 2.6M \\
                        & C2T & $\beta=0.3$ & 4.59 & 1.9M \\
\hline                        
\multirow{2}{*}{L2 70B} & E-2 & $N=260$ & 4.08 & 2.5M \\
                        & C2T & $\beta=0.3$ & 4.08 & 2.0M \\
\hline
\multirow{3}{*}{V 7B} & E-2 & $N=180$ & 5.25 & 1.5M \\
                      & C2T & $\beta=0.35$ & 5.30 & 1.3M \\
                      & C2T* & $\beta=0.25$ & 5.32 & 1.1M \\
\hline                      
\multirow{3}{*}{V 13B} & E-2 & $N=200$ & 4.02 & 1.4M \\
                       & C2T & $\beta=0.35$ & 4.03 & 1.2M \\
                       & C2T* & $\beta=0.25$ & 4.03 & 1.1M \\
\hline                       
\multirow{3}{*}{V 33B} & E-2 & $N=210$ & 3.70 & 2.0M \\
                       & C2T & $\beta=0.3$ & 3.72 & 2.0M \\
                       & C2T* & $\beta=0.20$ & 3.74 & 1.8M \\
\hline
\end{tabular}
\renewcommand{\arraystretch}{1}
\caption{Comparison of the inference performance of Eagle-2 and our naive method without second TopK pruning on different models using MT-bench. L2 represents LLaMA-2, V represents Vicuna, E-2 represents EAGLE-2, C2T represents our naive method using the Classifier trained on the token trees inferred by LLaMA-2 7B, and C2T* represents using the Classifier fine-tuned on the token trees inferred by Vicuna-7B. M represents a million.}
\label{tbl:model_trans}
\end{table}

\subsection{Model Transferability}
\label{sec:model trans}

We trained the classifier on LLaMA-2 7B's token tree and tested its transferability to other models (LLaMA-2 13B, 70B, Vicuna 7B, 13B, and 33B) by freezing its parameters. Results in Table \ref{tbl:model_trans} show that C2T performs well on LLaMA-2 models that stably utilizes 75\% to 80\% of $\gamma$ while keeping $\tau$ unchanged. But the ability has declined on Vicuna models, though still better than EAGLE-2. When the classifier is fine-tuned on Vicuna's token trees, its performance comes back to the level achieved on the LLaMA-2 model family. The cost of this fine-tuning is minimal. For details, please refer to Appendix \ref{apx:fine-tuning}.

In summary, C2T has strong transferability within the same model family to still obtain good performance but may require fine-tuning for optimal performance when transferring to a different model family.

\begin{figure*}[htbp]
\centering
\subfigure[Relationship Graph between Verify Tokens and Time]{
\begin{minipage}[t]{0.45\textwidth}
\centering
\includegraphics[width=\textwidth]{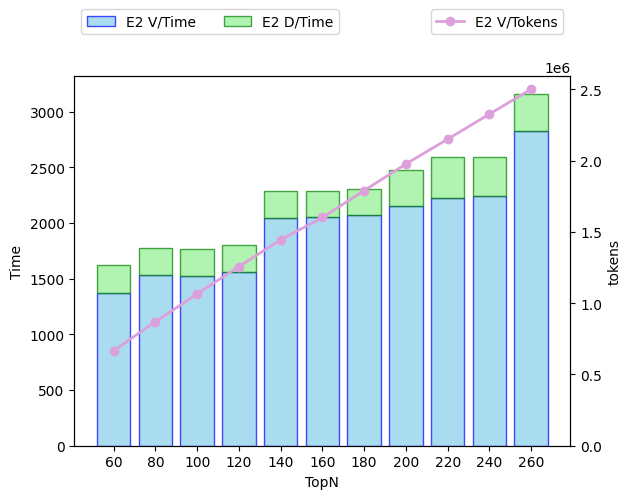}
\label{fig:e40}
\end{minipage}
}
\subfigure[Comparison Graph of EAGLE-2 and Ours]{
\begin{minipage}[t]{0.45\textwidth}
\centering
\includegraphics[width=\textwidth]{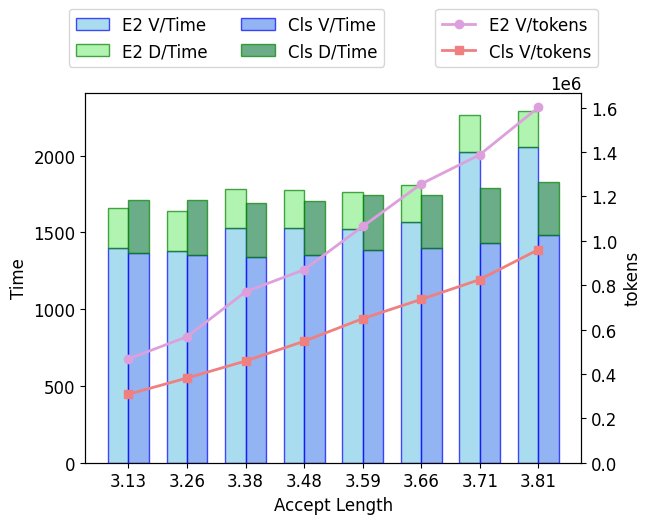}
\label{fig:e41}
\end{minipage}
}

\caption{The graph shows the relationship between candidate tokens number $\gamma$ and total wall-clock time on MT-bench using LLaMA-2 70B, with topK=15 and $d_{max}$=10. Total time is split into draft time (D/Time) and verify time (V/Time), shown as bar chart subparts on the left axis, while $\gamma$ (V/Tokens) are shown as a line chart on the right axis. The first figure uses topN as the x-axis, and the second figure uses accept length $\tau$ to align the methods.}
\label{fig:e4}
\end{figure*}

\subsection{Speed up in Larger LLMs}
\label{sec:bidmodel}

To evaluate the time efficiency of C2T under GPU limits, we tested the LLaMA-2 70B model on 2 * A100 (80G) GPUs. Figure \ref{fig:e40} shows that with EAGLE-2, as Top$N$ increases, $\gamma$ grows linearly but verify time increases in steps, not linearly.

We compared C2T with EAGLE-2 by matching $\tau$. Figure \ref{fig:e41} shows that C2T has a slightly higher draft time, but lower $\gamma$ resulting in a time advantage. When the parallel computing capability of GPUs is pushed to the limit, our method reduces the time by 18\% compared to EAGLE-2 under the same $\tau$. For detailed latency analysis, see Appendix \ref{apx:latency expriment}.

\subsection{Benefits in Chain Mode}
\label{sec:chain}

\begin{table}[H]
\centering
\renewcommand{\arraystretch}{1.25}
\begin{tabular}{c c c c c}
\hline
\textbf{Method} & \textbf{Avg length} & \textbf{$\tau$} & \textbf{$\gamma$} \\
\hline
\multirow{6}{*}{EAGLE 1/2}
                       & 5 & 1.98 & 112231 \\
                       & 6 & 2.06 & 125992 \\
                       & 7 & 2.10 & 140238 \\
                       & 8 & 2.09 & 155420 \\
                       & 9 & 2.13 & 184140 \\
\hline
DyMax & 6.10 & 2.06 & 127018 \\
DyJoint & 6.33 & 2.12 & 129876 \\
\hline
C2T & 5.46 & 2.12 & 116694 \\               
\hline
\end{tabular}
\renewcommand{\arraystretch}{1}
\caption{The comparison between other methods and C2T with $\beta=0.85$ in chain mode using the LLaMA-2 7B model on the MT-bench. DyMax and DyJoint represent the dynamic methods using the maximum probability with threshold=0.3 and joint probability with threshold=0.08 as the criterion for early stopping respectively. The maximum depth for all the dynamic methods is 10. Avg length represents the average generation length without the initial token.}
\label{tbl:chain}
\end{table}

Experiments in chain mode are meaningful because current dynamic tree construction does not support batch sizes greater than 1. This is due to the inability to have different attention masks within the same batch. In contrast, a chained token tree is always compatible with multi-batch scenarios.

In chain mode, C2T degrades as an early exit strategy and EAGLE-2 is rendered ineffective, essentially reverting to EAGLE-1. We varied the maximum draft length from 5 to 9 tokens for EAGLE-1/2. Additionally, we compared dynamic methods using the maximum probability and joint probability as early stopping criteria. The results, shown in Table \ref{tbl:chain}, demonstrate that C2T retains an advantage in chain mode.

\section{Conclusion}
In this paper, we propose C2T to address the limitations of previous tree construction methods that rely solely on joint probability. By training a classifier with additional features, we improved the precision of tree construction. We conducted extensive evaluations on multiple benchmarks and various LLMs to compare with the SOTA method EAGLE-1/2. Our method achieved superior results in all experiments and demonstrated strong transferability and applicability. C2T can construct a more precise tree using 75\% to 80\% of the tokens while maintaining the acceptance length. When the parallel computing capability of GPUs is pushed to the limit, C2T reduces the time by 18\% compared to EAGLE-2 under the same acceptance length.

\section*{Limitation}

C2T, similar to other dynamic tree construction approaches, currently does not support batch sizes greater than 1 (bs > 1) due to the use of different tree masks within the same batch, which is not supported by existing engineering implementations. However, C2T supports early stopping in chain mode, which is compatible with bs > 1. In practical industry use, bs > 1 is typically used in conjunction with chain mode \citet{liu2024deepseek}. And the preliminary combo experiments of the MTP-style layer in DeepSeek-V3 and C2T are in Appendix \ref{apx:mtp}.

Compared to methods using joint probability directly, C2T adds minimal overhead. This overhead is negligible when verification time is significant. And the quantitative analysis shows that the additional FLOPs are negligible and imply potential for optimization in engineering implementation. Please refer to Appendix \ref{apx:latency quant} for details.

\section*{Ethics Statement}

Our research adheres to the ACL Code of Ethics. We have ensured that our work respects user privacy and does not include any personal information in the datasets used. The datasets are publicly available and were labeled through interactions with English-speaking users. The tools and models used in this study are utilized in compliance with their intended purposes and are accessible under permissive licenses. We are committed to upholding the ethical standards of the ACL and promoting responsible research practices within the NLP community.



\bibliography{latex/reference}

\newpage
\appendix
\section{EAGLE-1 and EAGLE-2}
\label{apx:eagle1 and eagle2}

We conducted our experiments on LLaMA-2 7B and MT-bench. According to Table 1 of EAGLE-2 \citet{li2024eagle2}, the accept length of EAGLE-1 \citet{li2024eagle} is 3.62, and that of EAGLE-2 is 4.70. We also obtained similar results in our reproduction. However, during the experiment, we found that this comparison is not fair. Based on Appendix A of EAGLE-1 paper and EAGLE-2 paper, we can obtain the shapes of their respective token trees, where the size of the token tree in EAGLE-1 is 26, while that in EAGLE-2 is 60. Therefore, we further aligned the sizes of the token trees of both models and introduced our method C2T. The experimental results are shown in Figure \ref{tbl:eagle1 and eagle2}.

\begin{table}[H]
\centering
\renewcommand{\arraystretch}{1.25}
\begin{tabular}{c c c c}
\hline
\textbf{method} & \textbf{setting} & \textbf{$\tau$} & \textbf{$\gamma$} \\
\hline
E-1                  & Top$N$=26 & 2.66 & 340054 \\
\hline
\multirow{2}{*}{E-2} & Top$N$=26 & 2.95 & 317668 \\
                     & Top$N$=60 & 3.65 & 628980 \\
\hline
\multirow{2}{*}{C2T} & $\beta$=0.85 & 3.06 & 226569 \\
                     & $\beta$=0.65 & 3.66 & 531947 \\
\hline
\end{tabular}
\renewcommand{\arraystretch}{1}
\caption{The experiments on LLaMA-2 7B and MT-bench regarding the capability comparison between EAGLE-1, EAGLE-2, and C2T. TopN represents the size of the token tree, $\tau$ represents the accept length, and $\gamma$ represents the total number of candidate tokens to be verified. For EAGLE-2 and C2T, Top$K$=10 and $d_{max}$=6. It should be noted that our $\tau$ does not include the initial token generated by the target model, which is always accepted. Therefore, our $\tau$ is 1 smaller than that reported in the EAGLE-2 paper.}
\label{tbl:eagle1 and eagle2}
\end{table}

After aligning the tree sizes, EAGLE-2's performance falls short of expectations, with only an 11\% improvement in accept length. In practice, dynamic methods incur additional costs, leading to worse wall-clock times (on 2 A100 80G GPUs). While dynamic methods generate more accurate token trees than static methods, the extra computational cost means the tree size must be increased to achieve a speedup. Essentially, dynamic methods optimize GPU utilization further, as manually designing larger trees is extremely difficult and impractical for complex scenarios. However, given GPU limitations, increasing tree size also increases the verification burden on the target model. Thus, C2T's ability to generate more compact trees while maintaining the same accept length is particularly valuable.

\section{Classifier}
\label{apx:classifier}

\begin{figure*}[ht]
\centering
\begin{minipage}{0.33\textwidth}
\includegraphics[width=\textwidth]{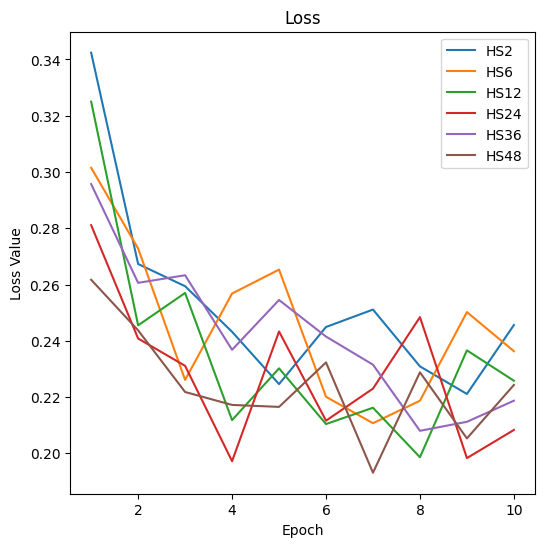}
\end{minipage}%
\begin{minipage}{0.33\textwidth}
\includegraphics[width=\textwidth]{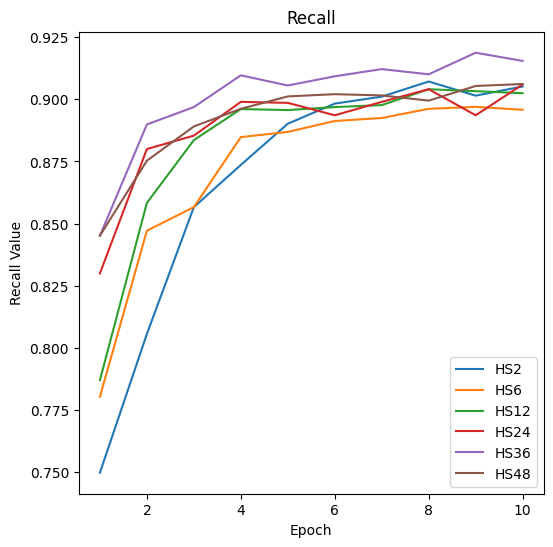}
\end{minipage}%
\begin{minipage}{0.33\textwidth}
\includegraphics[width=\textwidth]{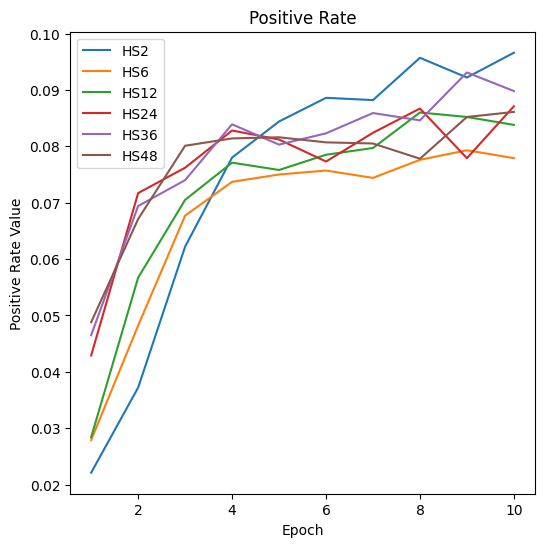}
\end{minipage}
\caption{The training process of FFNs with different structures, where HS$n$ represents a two-layer FFN with hidden state size $n$.}
\label{fig:e5}
\end{figure*}

\begin{figure*}[ht]
\centering
\begin{minipage}{0.33\textwidth}
\includegraphics[width=\textwidth]{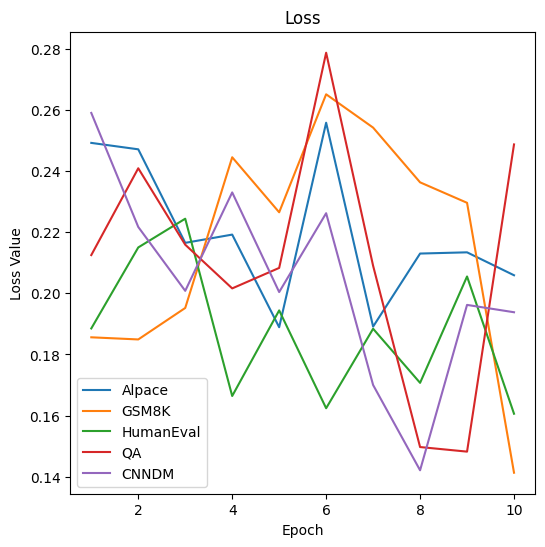}
\end{minipage}%
\begin{minipage}{0.33\textwidth}
\includegraphics[width=\textwidth]{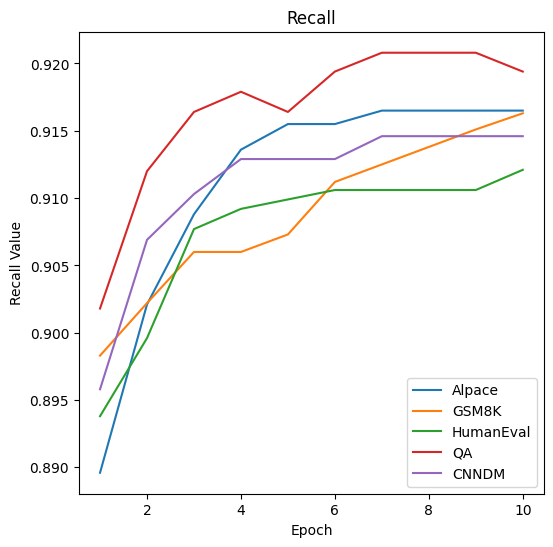}
\end{minipage}%
\begin{minipage}{0.33\textwidth}
\includegraphics[width=\textwidth]{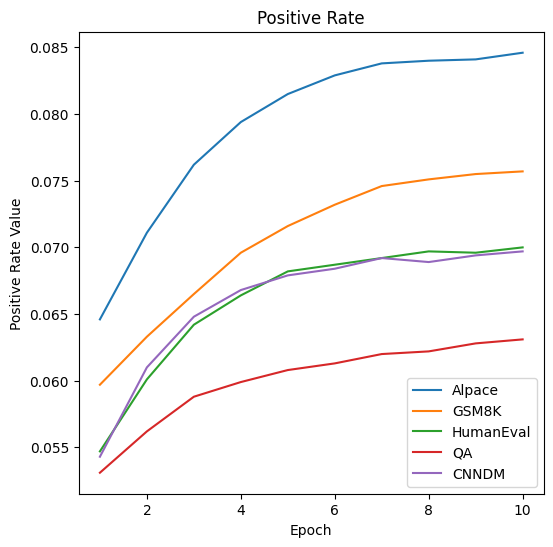}
\end{minipage}
\caption{The fine-tuning process of the classifier with the benchmark.}
\label{fig:e6}
\end{figure*}

\subsection{Dataset}

The classifier used in this paper, if not specifically mentioned, is trained on the token tree generated by LLaMA-2 7B on the MT-bench using the EAGLE-2 strategy. The settings are $d_{max}$ = 11 (excluding the root generated by $M_t$), Top$K$ = 10, and the Top$N$ = 1011 (meaning no recall is performed during the rerank stage, and the complete tree is used as the training dataset). Each data entry uses joint probability, depth, and entropy as features, and whether it is accepted as the label. We simply clean the data by dropping entries containing NA values, resulting in 8880 token trees, each containing 1011 nodes, for a total of 8880 * 1011 = 8,977,680 training data entries.

\subsection{Training and Evaluation}
\label{apx:C2T train and eval}

We split the dataset in a ratio of 0.95:0.05. Since the dataset has sparse positive samples, for example, each token tree has 1011 nodes, however, only 3.5 tokens are accepted by the $M_t$, so we perform negative sampling on this sparse dataset. During training, we set the batch size to 1024, and during evaluation, to align with the token tree verification configuration, we set the batch size to 1011. We use Adam as the optimizer with a learning rate (lr) of $1 \times 10^{-3}$ and train for 10 epochs, and use BCE as the criterion.

For evaluation, we focus more on recall, meaning the classifier should try to recall all tokens that are ultimately accepted by $M_t$. At the same time, we should also pay attention to the positive rate, which is the probability that the classifier predicts a token as positive, denoted as $\theta$. This value corresponds to the ratio of Top$N$ to the size of $T_1$ in EAGLE-2 and is positively correlated with the final $\gamma$. When selecting the classifier, priority should be given to models with significantly higher recall. Among models with similar recall, choose the one with a smaller $\theta$.

\subsection{Structure}

In this paper, we also briefly explored the effects of classifiers with different FFN structures. We mainly discussed the performance of FFNs with two layers and different hidden states, setting the hidden state of the classifier to 2, 6, 12, 24, 36, and 48, respectively. We trained various FFNs according to the training configuration in \ref{apx:C2T train and eval}, and the training process is shown in Figure \ref{fig:e5}:

Furthermore, we applied these classifiers to the C2T inference of LLaMA-2 7B on MT-bench, aligning the threshold $\beta = 0.5$, It can be observed that, with similar $\gamma$, the $\tau$ for the six classifier structures are 3.91, 3.97, 4.02, 4.03, 4.02, and 4.02, respectively. Therefore, it can be concluded that for our task, a hidden state of 12 to 48 is more appropriate for the classifier. In our other experiments, this value is set to 48 by default.

\subsection{Fine-tuning}
\label{apx:fine-tuning}

In this paper, we also explored the fine-tuning, using Adam as the optimizer with a lr of $1 \times 10^{-4}$ and training for 10 epochs using 10\% of the data. In addition to fine-tuning in the Experiment \ref{sec:model trans} testing model transferability, we also attempted fine-tuning between different token trees inferred on different benchmarks. The fine-tuning process is shown in Figure \ref{fig:e6}. As shown in the figure, although fine-tuning improves the classifier's recall, it also increases the positive rate. When we applied the fine-tuned classifier to inference, we found that the distribution relationship between candidate tokens number $\gamma$ and accept length $\tau$ remained almost unchanged. The only difference is that the fine-tuned classifier requires a larger $\beta$ to achieve the same $\gamma$ and $\tau$ as before. This means that the fine-tuned classifier becomes more confident, but there is no significant improvement in precision. This also indirectly proves the data-free characteristic of our classifier.

\subsection{Cross Validation}
\label{apx:cross validation}

\begin{figure}[ht]
\centering
\includegraphics[width=0.5\textwidth]{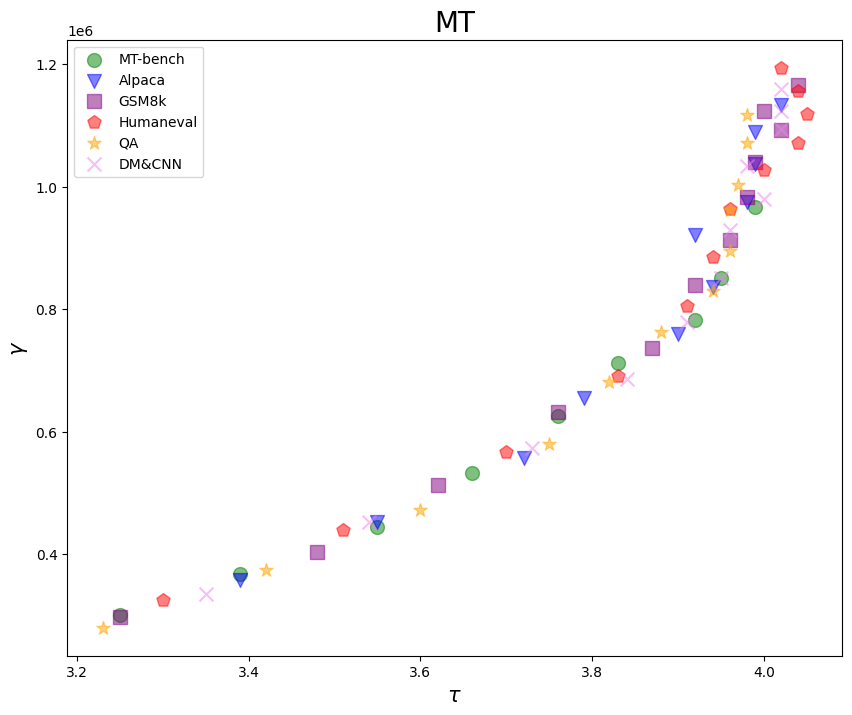}
\caption{Scatter plots of candidate tokens to $\gamma$ and accept length $\tau$ obtained by C2T on MT-bench using classifiers trained on different datasets.}
\label{fig:e9}
\end{figure}

Since our previous experiments involved inferring token trees on MT-bench to train the classifier, and then transferring the classifier to other datasets for speculative decoding, we now cross-validate the effectiveness of C2T on MT-bench. To do this, we train the classifier from scratch using token trees generated from other datasets and then apply it to inference on MT-bench. This experiment was conducted on LLaMA-2 7B. The results are shown in Figure \ref{fig:e9}. Classifiers trained on other datasets and those trained directly on MT-bench show nearly identical distributions in the scatter plots when used for inference with C2T. This cross-validates the feasibility of C2T on MT-bench.

\subsection{Use TopK for Second Pruning}
\label{sec:topk}

\begin{table}
\centering
\renewcommand{\arraystretch}{1.25}
\begin{tabular}{@{\hspace{10pt}}c@{\hspace{10pt}}@{\hspace{10pt}}c@{\hspace{10pt}}@{\hspace{10pt}}c@{\hspace{10pt}}@{\hspace{10pt}}c@{\hspace{10pt}}}
\hline
\textbf{$\beta$} & \textbf{$K$} & \textbf{$\tau$} & \textbf{$\gamma$} \\
\hline
\multirow{5}{*}{0.3} & 15 & 3.98 & 1026454 \\
                     & 20 & 4.15 & 1235635 \\                     
                     & 25 & 4.21 & 1413190 \\                      
                     & 30 & 4.30 & 1568173 \\                     
                     & $/$ & 4.41 & 2030064 \\                    
\hline
\multirow{5}{*}{    0.5    } & 15 & 3.92 & 781806 \\
                     & 20 & 4.01 & 878722 \\
                     & 25 & 4.02 & 931429 \\
                     & 30 & 4.02 & 971486 \\
                     & $/$ & 4.02 & 983491\\
\hline
\end{tabular}
\renewcommand{\arraystretch}{1}
\caption{The topK secondary pruning experiment of LLaMA-2 7B on MT-bench, where K represents the value of topK, $/$ represents the naive method without secondary pruning.}
\label{tbl:topk prime}
\end{table}

In the methodology section, we discussed the improvement of further constraining the tree shape using Top$K$ in pre-pruning. In the methodology, we mentioned that using Top$K$ to simplify the computation process is necessary, but the latter Top$K$ pruning after obtaining the confidence scores is optional. In this experiment, we will conduct a variable analysis for the latter case. This approach results in a smaller and more stable token tree, and by varying Top$K$, we can generate multiple scatter plots. As shown in \ref{fig:e3}, it is observed that after introducing Top$K$ for second pruning, the method maintains a similar distribution on the graph as the original method, and in some threshold values, it even yields slightly better results. Regardless of fluctuations, it consistently outperforms EAGLE-2. As shown in Table \ref{tbl:topk prime}, under a high $\beta$, secondary pruning can get gains in $\gamma$ with minimal loss of $\tau$, but it also limits the maximum capability under a low $\beta$. In our other experiments, we default to using Top$K$=15 for two complete rounds of constrained pruning.

\section{Proof of Simplified Calculation}
\label{apx:proof}
The method proposed in this paper, when calculating entropy, requires first obtaining the top$M$ probabilities and then calculating the entropy of these $M$ probabilities. Let the vocabulary size be $V$, and it is known that the implementation of torch.topk is based on the quickselect algorithm. 

If we consider only the calculation of entropy, obtaining the top$M$ probabilities first and then calculating the entropy is more complex than directly calculating the entropy. 

\begin{proof}[\textbf{Proof-1}]
    The FLOPs for directly calculating the entropy is {\small $F_1=2*V$}. In contrast, obtaining the top$M$ probabilities involves {\small $F_2=C_1*V$}, after calculating the entropy of $M$ probabilities involves {\small $F_3=F_2+2*M=C_1*V+2*M$}, where {\small $C_1>2$} in most cases. So {\small $2*V < C_1*V+2*M$}, which means {\small $F_1<F_3$}. In summary, selecting first and then calculating the entropy is more complex than directly calculating the entropy in most cases.
\end{proof}

However, we need to take into account the impact of the joint probability calculation step. Both C2T and EAGLE-2 only need to calculate the joint probabilities of the Top$K$ when computing joint probabilities. We first argue the necessity of this step.

\begin{proof}[\textbf{Proof-2}]
    If we were to fully calculate the joint probabilities and then select, since each tree layer has at most $K$ nodes, considering the parallel computing capability of GPUs, the overall complexity for calculating the joint probabilities is {\small $O(V)$}. There would be {\small $K * V$} probabilities in total, and selecting the Top$K$ from them would involve {\small $O(K * V)$}. Therefore, the total complexity would be {\small $O(V + K*V)$}. In contrast, by only taking the Top$K$ for each node, due to the parallel computing nature of GPUs, the total complexity for selection is {\small$O(V)$}, resulting in $K^2$ probabilities. The complexity for selecting the Top$K$ from these probabilities is {\small $O(K^2)$}, so the overall complexity is {\small $O(V + K^2)$}. Since {\small $K^2 << V$}, therefore {\small $O(V+K*V) > O(V + K^2)$}. In summary, it is necessary to first select and then calculate the joint probabilities.
\end{proof}

Therefore, what we are actually comparing are the complexities of the following two scenarios:
\begin{itemize}
    \item\label{sce1} Directly calculating the entropy and then selecting the Top$K$ probabilities.
    \item\label{sce2} First selecting the top$M$ probabilities, then calculating the entropy of these $M$ probabilities, and finally selecting the Top$K$ probabilities from these $M$ probabilities.
\end{itemize}

\begin{proof}[\textbf{Proof-3}]
    From \textit{\textbf{Proof-1}}, we know that for the first scenario, the FLOPs before taking the Top$K$ is {\small $F_1=2*V$}, and the FLOPs for taking the Top$K$ is {\small $F_4=C_2*V$}. Therefore, the total FLOPs is {\small $F_5=(2+C_2)*V$}. For the second scenario, the FLOPs before taking the Top$K$ is {\small $F_3=C_1*V+2*M$}, and the FLOPs for taking the Top$K$ from the $M$ probabilities is {\small $F_5=C_3*M$}. Therefore, the total FLOPs is {\small $F_6=C_1*V+(2+C_3)*M$}. Since for the quickselect algorithm, the final number of computations is independent of the number of elements to be selected, {\small $C_1 \approx C_2$}. Also, since {\small $V>>M$}, we have {\small $F_4-F_6 =(C_2-C_1)*V + 2*V-(2+C_3)*M \approx 2 * V-(2+C_3)*M>0$}, which means {\small $F_4>F_6$}. In summary, considering the computation of joint probabilities, the FLOPs of the first scenario are more than the second scenario.
\end{proof}

We have demonstrated the necessity of simplifying the calculation of entropy, and \textit{\textbf{Proof-3}} implies that, from the perspective of reducing FLOPs, $M$ should be as small as possible. However, an excessively small $M$ may lead to the long-tail effect. Therefore, we conducted experiments on $M$ using LLaMA-2 7B on the MT-bench with $\beta=0.5, topK=15$, and the results are shown in Table \ref{tbl:topM prime}. The results indicate that when $M=1000$, the impact of the long-tail effect is almost completely eliminated.

\begin{table}[H]
\centering
\renewcommand{\arraystretch}{1.25}
\begin{tabular}{c c c}
\hline
\textbf{$M$} & \textbf{$\tau$} & \textbf{$\gamma$} \\
\hline
100 & 3.70 & 815365 \\
500 & 3.89 & 796981 \\
1000 & 3.92 & 781613 \\
10000 & 3.92 & 782104 \\
$/$ & 3.92 & 781806 \\
\hline
\end{tabular}
\renewcommand{\arraystretch}{1}
\caption{The experiments on the values of $M$ using C2T with LLaMA-2 7B on MT-bench, with $\beta=0.5, topK=15$, where $/$ indicates no use of top$M$ for simplified calculation.}
\label{tbl:topM prime}
\end{table}

\section{Confidence}
\label{apx:confidence}

\begin{figure}[ht]
\centering
\includegraphics[width=0.5\textwidth]{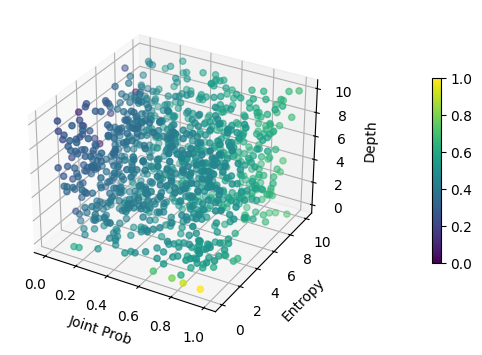}
\caption{Input-output 3D-heatmap, where the color becomes lighter as the outputs increase.}
\label{fig:input-output 3d-heatmap}
\end{figure}

\begin{figure*}[ht]
\centering
\subfigure[]{
\begin{minipage}[t]{0.315\textwidth}
\centering
\includegraphics[width=\textwidth]{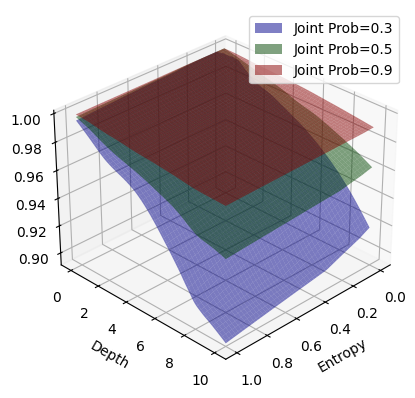}
\label{fig:e71}
\end{minipage}
}
\subfigure[]{
\begin{minipage}[t]{0.315\textwidth}
\centering
\includegraphics[width=\textwidth]{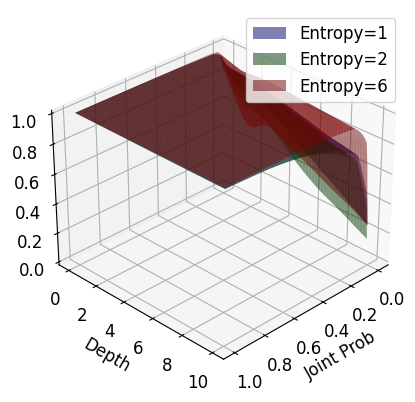}
\label{fig:e72}
\end{minipage}
}
\subfigure[]{
\begin{minipage}[t]{0.315\textwidth}
\centering
\includegraphics[width=\textwidth]{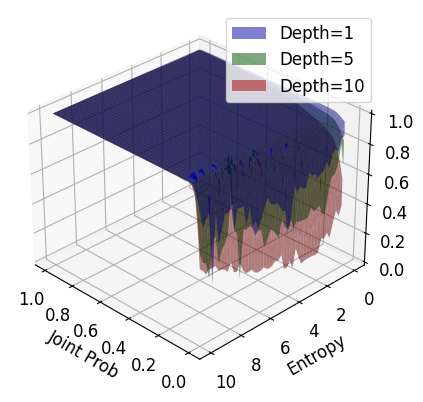}
\label{fig:e73}
\end{minipage}
}
\caption{Interpolation plots of input-output with some parameters fixed, where the z-axis represents the outputs of the classifier.}
\label{fig:e7}
\end{figure*}

Since the features of our classifier are a triplet and the output is a scalar, it is highly suitable for statistical analysis. First, we created a 3D heatmap for the overall input-output by randomly generating 1000 data points, where the joint probability is a float in [0,1], entropy is a float in [0,10], and depth is an integer in [0,10]. To highlight the outputs, the output is not normalized using sigmoid but rather using min-max normalization. The experimental results are shown in Figure \ref{fig:input-output 3d-heatmap}. Among these three features, the joint probability has the greatest impact on the confidence score, and this fitting result aligns with our intuition. The impact of entropy and depth is not as easily captured, hence we conducted further quantitative analysis.

Furthermore, we fixed one parameter and varied the other two to generate 10,000 data points. The interpolated results are shown in Figure \ref{fig:e7}. Figure \ref{fig:e71} shows that the three surfaces are widely spaced, indicating the significant impact of joint probability on the confidence score. The steeper slope with lower joint probability suggests greater influence from the other features. Figure \ref{fig:e72} shows that surfaces are closely adjacent at low entropy (1 or 2), indicating minimal impact on confidence scores. However, at high entropy, surfaces rise above those with lower entropy at low joint probability, showing positive confidence corrections for less confident distributions, addressing the case in Figure \ref{fig:shortage1}. Figure \ref{fig:e73} shows that surfaces coincide and flatten at high joint probability, indicating minimal effect of depth. Conversely, at low joint probability, surfaces steepen and diverge, with shallower depths yielding higher confidence scores, providing additional opportunities for subsequent node generation, addressing the case in Figure \ref{fig:shortage2}.

\begin{figure*}[ht]
\centering
\subfigure[EAGLE-2 Draft Latency]{
\begin{minipage}[t]{0.82\textwidth}
\centering
\includegraphics[width=\textwidth]{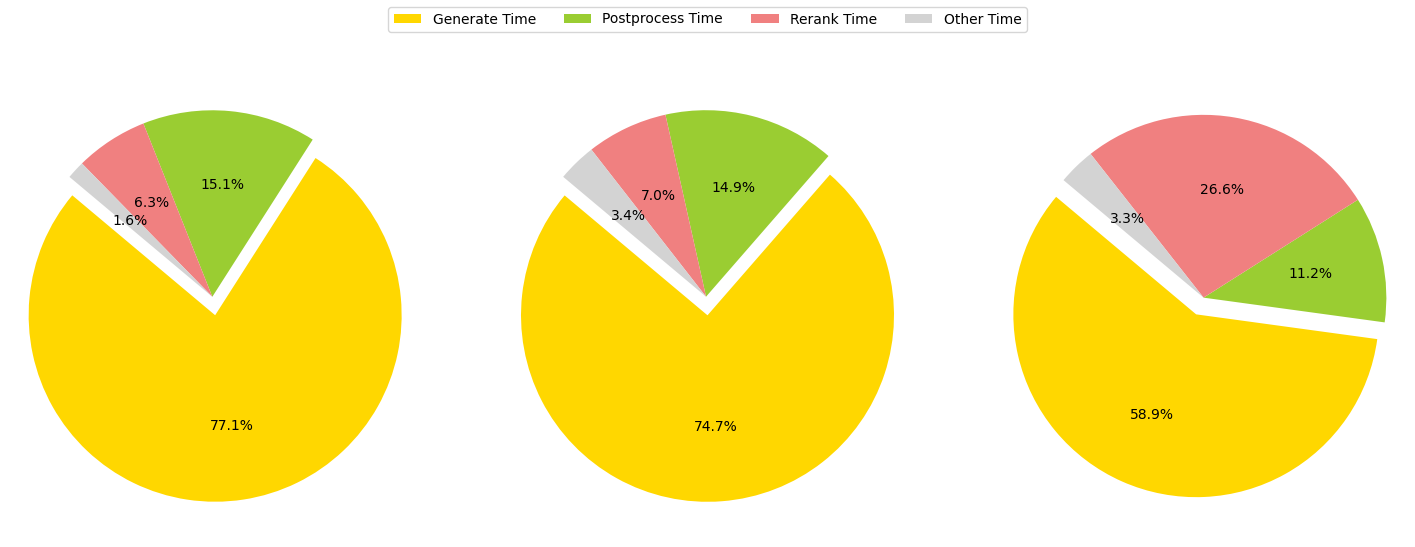}
\label{fig:e80}
\end{minipage}
}
\subfigure[Ours Draft Latency]{
\begin{minipage}[t]{0.82\textwidth}
\centering
\includegraphics[width=\textwidth]{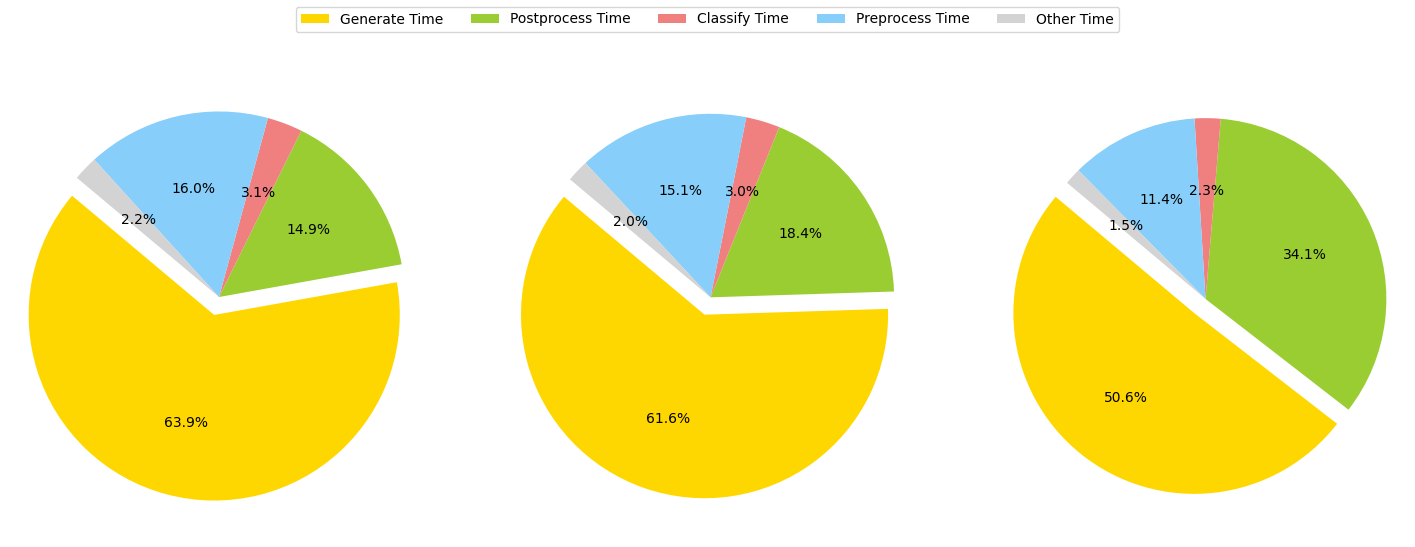}
\label{fig:e81}
\end{minipage}
}
\caption{Draft latency analysis using the LLaMA-2 model family on MT-bench. The top row is for EAGLE-2, and the bottom row is for Ours. From left to right are the results on LLaMA-2 7B, 13B, and 70B, respectively.}
\label{fig:e8}
\end{figure*}

\section{Additional Overhead}
\label{apx:latency}

\subsection{Experimental Analysis}
\label{apx:latency expriment}

Speculative decoding can be divided into two stages: Draft and Verify. Our code is modified based on EAGLE, without changing the Verify part, but only optimizing the Draft part. Therefore, we mainly focus on the latency analysis of the Draft stage. In C2T, for the Draft stage, we can further divide it into several cyclical sub-stages: 

\begin{enumerate}
    \item \textbf{Classify:} The classifier accepts input features and makes predictions.
    \item \textbf{Pre-processing:} Process the predictions of the classifier and obtain the input for the draft model.
    \item \textbf{Generate:} The draft model accepts input and generates the tokens.
    \item \textbf{Post-processing:} Process the output of the draft model and obtain the input features for the classifier.
\end{enumerate}

For EAGLE-2, we can align the granularity and divide its Draft into three stages:

\begin{enumerate}
    \item \textbf{Generate:} The draft model accepts input and produces output.
    \item \textbf{Post-processing:} Process the output of the draft model and obtain the input for the next step.
    \item \textbf{Rerank:} rerank the generated token tree based on joint probability and recall the Top$N$ nodes.
\end{enumerate}

We conducted latency analysis for the Draft stage of the LLaMA-2 7B, 13B, and 70B models on 2 * A100 (80G) GPUs to infer MT-bench using C2T with $topK=15$ for secondary pruning and a threshold $\beta=0.5$ and EAGLE-2 with $topN=80$, as shown in Figure \ref{fig:e8}. 

From Figure \ref{fig:e80}, it can be seen that in EAGLE-2, the proportion of Rerank Time significantly increases with the growth of model size. This is because when the GPU's computational capacity is pushed to its limit, the more complex rerank-recall operations also take longer to complete. From Figure \ref{fig:e81}, in C2T, the proportion of Postprocess Time significantly increases with the model size. Compared to EAGLE-2's Postprocess Time, we need to additionally calculate entropy and depth, since the latter can be obtained directly, the main impact is caused by the former. Moreover, to prevent the high complexity of calculating entropy due to a large vocabulary, we decompose the topK operation into two stages: top1000 and then topK. We use the 1000 probabilities selected in the first stage to calculate entropy, which introduces additional latency.

Since C2T and EAGLE-1/2 share the same draft model, the Generate Time is aligned between the two methods. The proportion of this item in each method indicates that C2T's Draft Time is slightly greater than EAGLE-2, but this gap is gradually decreasing with the increase in model size.

\subsection{Quantitative Analysis}
\label{apx:latency quant}

From the perspective of quantitative analysis, the additional computational overhead introduced by C2T compared to EAGLE-2 primarily consists of the calculation of entropy, the computation of depth, and the output of the classifier. Since the computation of joint probability is inherently coupled with the EAGLE-2's generation process, and depth can be directly obtained during the construction of the tree structure, both can be considered negligible. Therefore, the focus of our analysis is on the computational complexity of entropy calculation and the forward pass of the classifier. In addition, when calculating the entropy, we first select the top$M$ values before computing the entropy. Similarly, for EAGLE-2, the engineering implementation of calculating joint probability also requires selecting the Top$K$ values first. According to Appendix \ref{apx:proof} \textit{\textbf{Proof-3}}, the costs of these two parts can offset each other. Therefore, we only need to consider the complexity introduced by calculating the entropy over $M$ values which is $O(M)$.

Then, considering the worst-case scenario, where $K$ tokens participate in tree construction at each tree layer with a maximum depth of $d_{max}$, the total complexity for entropy calculation is $O(K*d_{max}*M)$.

Given that our classifier consists of two layers with a hidden layer size of $h$:

\begin{itemize}
    \item the complexity for matrix multiplication from the input to the hidden layer is $O(3h)$.
    \item the activation function computation in the hidden layer is $O(h)$.
    \item the matrix multiplication from the hidden layer to the output layer is $O(h)$
\end{itemize}
 
Under the worst-case scenario, the total classification complexity is $O(K*d_{max}*5h)$.

In summary, the overall additional complexity introduced by C2T is $O(K*d_{max}*(M+5h))$.

In practical scenarios, with $M=1000$, $K=15$, $d_{max}=10$, $h=48$, the complexity for entropy calculation is $O(150,000)$, and the complexity for the classifier is $O(36,000)$. The total additional overhead complexity is $O(186,000)$.

This complexity is almost negligible compared to the complexity of a single forward pass of LLMs. Compared with our experimental results, it implies a great space for engineering optimization.

\section{Combo with MTP-style Layer}
\label{apx:mtp}

In the latest industry application of speculative decoding, DeepSeek-V3 \citet{liu2024deepseek}, inspired by the training method of EAGLE's draft model, a Multi-Token Prediction (MTP) approach is proposed. Specifically, based on EAGLE, the embeddings of input tokens and the hidden states are firstly normalized by RMSNorm \citet{zhang2019root}, then concatenated, and finally dimensionally reduced through the linear head. We froze the parameters of the previously trained classifier and directly transferred it to the inference of the EAGLE layer trained in an MTP-style manner. Preliminary experiment is in Table \ref{tbl:mtp combo}. 

\begin{table}[H]
\centering
\renewcommand{\arraystretch}{1.25}
\begin{tabular}{c c c c c}
\hline
\textbf{model} & \textbf{strategy} & \textbf{setting} & \textbf{$\tau$} & \textbf{$\gamma$} \\
\hline
\multirow{4}{*}{EAGLE} & C & $/$ & 1.96 & 161020 \\
                       & C + C2T & $\beta$=0.9 & 1.97 & 104141 \\
                       & E-2 & Top$N$=90 & 3.68 & 935280 \\
                       & C2T & $\beta$=0.5 & 3.70 & 800848 \\
\hline
\multirow{4}{*}{MTP} & C & $/$ & 2.12 & 156040 \\
                     & C + C2T & $\beta$=0.8 & 2.10 & 98141 \\
                     & E-2 & Top$N$=90 & 3.86 & 900090 \\
                     & C2T & $\beta$=0.5 & 3.84 & 724913 \\               
\hline
\end{tabular}
\renewcommand{\arraystretch}{1}
\caption{The MTP combo experiments on LLaMA-2 7B and MT-bench. C represents chain-like drafting, C + C2T represents using C2T to early stopping in chain-like drafting, E-2 represent EAGLE-2. The default parameters are Top$K$=15 and $d_{max}$=10.}
\label{tbl:mtp combo}
\end{table}

It should be noted that, in order to align with the MTP experimental configuration, the EAGLE layer here is retrained by us, hence there is some data deviation from previous experiments that directly used the model provided by EAGLE.

\end{document}